\documentclass[letterpaper, 10 pt, conference]{ieeeconf}  

\IEEEoverridecommandlockouts                              

\let\labelindent\relax

\usepackage{macros}

\begin{document}

\title{\LARGE \bf Low-Cost Underwater In-Pipe Centering and Inspection\\Using a Minimal-Sensing Robot}

\author{Kalvik Jakkala and Jason O'Kane
\thanks{The authors are with the Computer Science Department, Texas A\&M University, College Station, TX, USA. Email:{\tt\small \{kalvik, jokane\}@tamu.edu}}%
}

\maketitle
\thispagestyle{empty}
\pagestyle{empty}

\maketitle

\begin{abstract}
Autonomous underwater inspection of submerged pipelines is challenging due to confined geometries, turbidity, and the scarcity of reliable localization cues. This paper presents a minimal-sensing strategy that enables a free-swimming underwater robot to center itself and traverse a flooded pipe of known radius using only an IMU, a pressure sensor, and two sonars: a downward-facing single-beam sonar and a rotating 360° sonar. We introduce a computationally efficient method for extracting range estimates from single-beam sonar intensity data, enabling reliable wall detection in noisy and reverberant conditions. A closed-form geometric model leverages the two sonar ranges to estimate the pipe center, and an adaptive, confidence-weighted proportional–derivative~(PD) controller maintains alignment during traversal. The system requires no Doppler velocity log, external tracking, or complex multi-sensor arrays. Experiments in a submerged 46~cm-diameter pipe using a Blue Robotics BlueROV2 heavy remotely operated vehicle demonstrate stable centering and successful full-pipe traversal despite ambient flow and structural deformations. These results show that reliable in-pipe navigation and inspection can be achieved with a lightweight, computationally efficient sensing and processing architecture, advancing the practicality of autonomous underwater inspection in confined environments.
\end{abstract}

\IEEEpeerreviewmaketitle

\section{Introduction}

Submerged pipelines—including levees, culverts, and oil pipelines—form the backbone of municipal water networks, energy infrastructure, and industrial facilities. Routine inspection is essential to identify structural degradation, leaks, or blockages before they escalate into costly failures. Yet manual inspection is rarely feasible: pipes are typically narrow, flooded or pressurized, extend for hundreds of meters, and may contain hazardous or oxygen-depleted environments. Entering such confined spaces poses safety risks and is often impossible for humans. These challenges underscore the need for robotic systems capable of autonomously navigating and collecting inspection data inside pipes.

Despite the clear need, flooded pipe inspection remains a formidable challenge for robotics. Confined geometries restrict maneuverability, the absence of GPS precludes absolute localization, and turbidity or biofouling severely limits the effectiveness of cameras as primary sensors. These constraints have motivated the development of robotic platforms that rely on acoustic sensing and autonomous control to robustly maintain position within pipes.

Existing approaches span a wide spectrum, from crawler-type robots that press against the pipe walls~\cite{TurG10, MateosDV13, KakogawaSH14} to free-swimming vehicles equipped with multiple sonars and a Doppler velocity logger~(DVL)~\cite{MazumdarLFH12, WuNKYB15}. While effective in certain settings, these systems often involve complex mechanical designs, dense sensor payloads, or reliance on external infrastructure—factors that increase cost and operational burden. Vision-based methods~\cite{NarimaniNL09, ManjunathaSGM18} can reconstruct pipeline geometry in clear water, but their performance rapidly degrades under turbidity. Sonar-based centering algorithms~\cite{KazeminasabSSJRZB21, SewellPS24} offer a promising alternative, though many depend on multi-sensor arrays that are computationally demanding or impractical for compact vehicles.

In this work, we present a lightweight, sonar-based perception and control framework that enables an autonomous underwater robot to remain centered within a submerged pipe using only an IMU, a pressure sensor, and two sonar measurements: a downward-facing single-beam sonar and a 360° mechanically rotating single-beam sonar. 

\begin{figure}[t]
    \centering
    \includegraphics[width=0.75\linewidth]{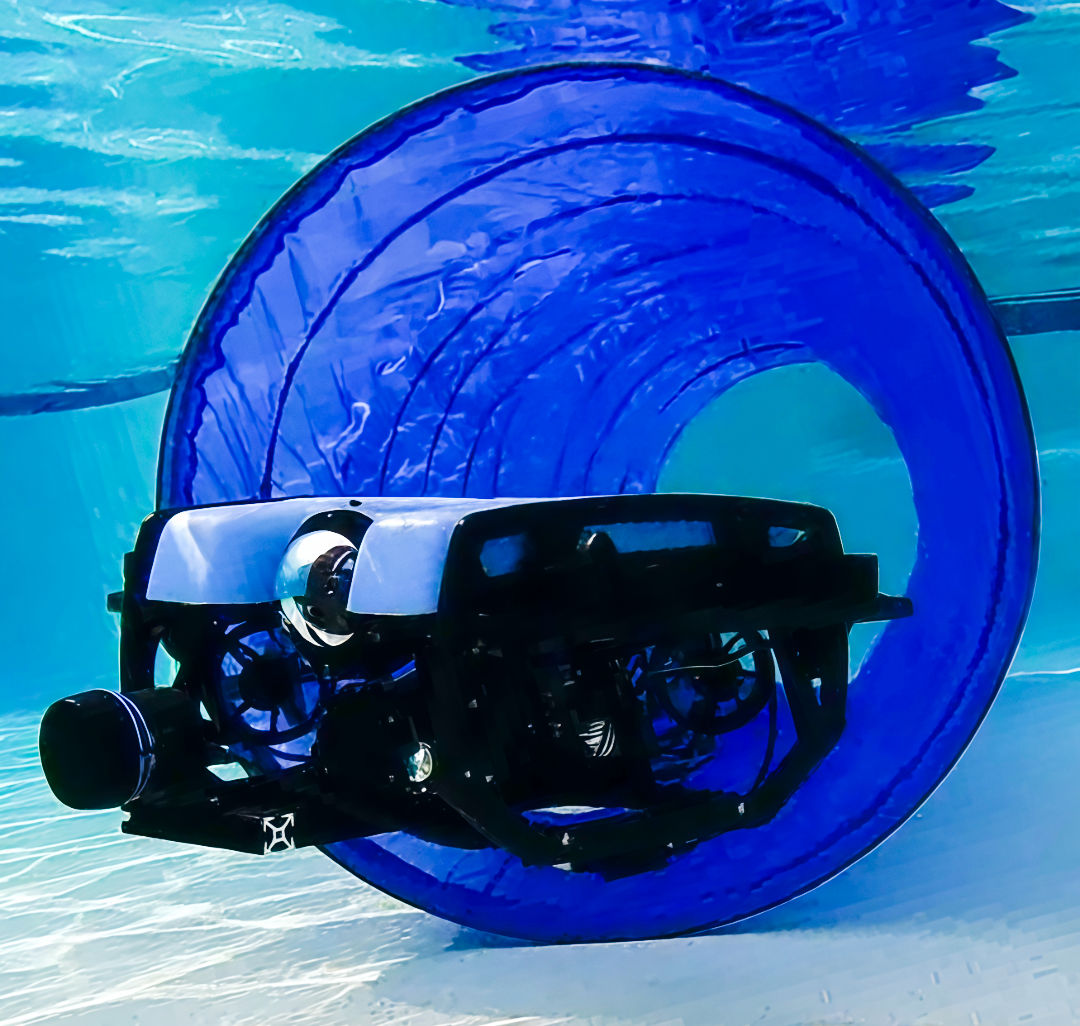}
    \caption{The Blue Robotics BlueROV2 Heavy vehicle used for experiments, equipped with a Ping360 rotating sonar mounted on the nose for 360° scanning and a Ping1D downward-facing sonar for bottom range measurement.}
    \label{fig:BREV}
\end{figure}

A key contribution of this work is the development of an efficient method for range estimation from raw single-beam sonar data. Unlike single-beam altimeter sonars that output range measurements explicitly, mechanically scanning sonars provide only acoustic intensity profiles for each azimuth angle. Extracting reliable wall ranges from these profiles is difficult due to near-field artifacts, multipath interference, and low signal-to-noise ratios in underwater environments. Conventional approaches often rely on static amplitude thresholds to detect wall echoes~\cite{Nielsen91}, which can be highly sensitive to noise and environmental variability. More recent work, such as Yang~et~al.~\cite{YangLRL23}, applies deep learning to interpret raw acoustic data and can achieve good performance, but at substantial computational cost—rendering these methods unsuitable for real-time operation on robotic platforms.

To overcome these limitations, we propose a computationally efficient method that extracts the first valid wall return from each intensity profile using signal processing methods. The approach provides consistent range estimates with minimal computational overhead, enabling real-time operation on resource-constrained robots. This efficient sonar processing pipeline serves as the foundation for the proposed geometric centering algorithm and facilitates simultaneous navigation and inspection using minimal hardware.

The pipe-centering method assumes the pipe radius is known a priori—a realistic condition for engineered infrastructure—and employs a closed-form geometric model that uses the two sonar ranges to estimate the robot’s offset from the pipe center. A confidence-weighted proportional–derivative~(PD) controller then regulates the robot’s motion to minimize lateral and vertical displacement, enabling stable traversal even in the presence of pipe irregularities and deformations.

We validate the full framework through field experiments in a submerged, flexible pipe testbed using a Blue Robotics BlueROV2 heavy remotely operated vehicle. The robot consistently maintained stable centering across varying initial offsets and pipe deformations, demonstrating that stable in-pipe navigation can be achieved using low-cost hardware and efficient acoustic processing.

The main contributions of this paper are as follows:
\begin{enumerate}
    \item A computationally efficient algorithm for extracting range data from raw single-beam sonar profiles.
    \item A closed-form geometric formulation for estimating the pipe center using a minimal sensor configuration and a known pipe radius.
    \item Experimental validation of stable centering and traversal in a submerged flexible pipe using a low-cost robotic platform.
\end{enumerate}

\section{Related Work}

Research on robotic inspection in confined structures spans several domains, including underground tunnels, ground robotics in constrained passages, and in-pipe inspection. Surveys of in-pipe systems~\cite{TurG10, KazeminasabSSJRZB21} review a wide range of locomotion, sensing, and inspection strategies for civil infrastructure, while Montero et al.~\cite{MonteroVMJB15} and Tardioli et al.~\cite{TardioliRSRLVM19} synthesize lessons from a decade of tunnel and ground robotics research. Although these works provide valuable perspectives, underwater in-pipe inspection remains comparatively underexplored, despite sharing core challenges such as localization in GPS-denied environments, maintaining stability in narrow passages, and achieving effective sensing with minimal payloads.

Within in-pipe robotics more generally, many works have emphasized mechanical design and centering. Mateos et al.~\cite{MateosDV13} developed a control framework for automatic in-pipe centering by reducing 3D dynamics to a 2D controller, while Kakogawa et al.~\cite{KakogawaSH14} proposed underactuated crawler modules for confined locomotion. Tan et al.~\cite{TanSAWF19} investigated sparse sensor arrays for aerial robots navigating hazardous tunnels. Chen and Fang~\cite{ChenF23} analyzed optimal robot designs for generic pipe operations, and Patel et al.~\cite{PatelAATOASAAFAHJS24} introduced a multi-robot system for cooperative shallow-water pipeline inspection. On the control side, Sewell et al.~\cite{SewellPS24} demonstrated a nonlinear Proportional-Integral-Derivative~(PID) approach using sonar feedback, showing that acoustic ranges can regulate vehicle centering in the absence of complex localization.

In contrast, underwater in-pipe and subsea pipeline inspection has focused more directly on sensing for navigation and damage detection. Vision-based trackers~\cite{NarimaniNL09, ManjunathaSGM18} have been developed for pipeline inspection, but their performance is severely limited in turbid water. Acoustic methods offer greater robustness: Mazumdar et al.~\cite{MazumdarLFH12} designed a compact robot for nuclear piping inspection, Wu et al.~\cite{WuNKYB15} built a maneuverable swimmer for submerged pipe missions, and Shi et al.~\cite{ShiCGGHPXST17} demonstrated a sonar-guided tracking system for amphibious spherical robots. While these works validate sonar as a primary sensing modality for confined underwater environments, they often depend on multiple sonars, side-scan sensors, or Doppler velocity loggers~(DVLs).

Our work differs in that it tackles underwater in-pipe inspection using an extremely minimal sensor suite: an IMU, a pressure sensor, a downward-facing single-beam sonar, and a mechanically rotating 360° single-beam sonar. We show that real-time centering and inspection are achievable using only these low-cost sensors. This positions our contribution at the intersection of underwater inspection and minimal-sensing control, advancing in-pipe robotics toward practical, deployable systems for submerged environments.
\section{Problem Statement and System Description}

This work investigates the autonomous traversal of a submerged culvert, modeled as a straight cylindrical pipe of known radius $r$. The primary objective is for the robot to navigate safely along the pipe’s central axis using onboard sensors, while simultaneously collecting sonar data to assess the structural integrity of the pipe.

We employ a Blue Robotics BlueROV2 heavy remotely operated vehicle~(ROV; Figure~\ref{fig:BREV}) as the robotic platform. The vehicle is holonomic, providing full six-degrees-of-freedom via eight individually actuated thrusters, each with a 7.6 cm–diameter propeller.

The ROV is equipped with the following onboard sensors:

\begin{itemize}
    \item \textbf{360° Rotating Single-Beam Sonar:} A nose-mounted Blue Robotics Ping360 mechanical scanning sonar provides cross-sectional measurements of the pipe interior. At each azimuth angle, it emits an acoustic ping and records echo intensity as a function of time, producing a one-dimensional profile of reflected energy at 15 Hz. As the sensor mechanically rotates, these profiles are collected sequentially, requiring approximately $4.5$ s to complete a full 360° scan of the environment.
    
    \item \textbf{Downward-Facing Single-Beam Sonar:} A Blue Robotics Ping1D sonar mounted beneath the ROV measures the range to the pipe’s lower surface at 30~Hz. Unlike the 360° sonar, the Ping1D directly provides a processed range measurement. This is feasible because, unlike the 360° sonar, the Ping1D consistently observes the same downward target and operates at a higher update rate, enabling relatively reliable range estimation.
    
    \item \textbf{Pressure-Based Depth Sensor:} An onboard barometric pressure sensor measures absolute water pressure at 30~Hz and converts it to depth using a standard fluid statics model. Although it cannot provide spatially resolved distance relative to the pipe, it offers a stable, drift-free vertical reference that helps regulate the robot’s motion inside the pipe.

    \item \textbf{Inertial Measurement Unit (IMU):} A 30~Hz IMU provides tri-axial acceleration, angular velocity, and magnetic field measurements. Although the onboard magnetometer is included, its readings are unreliable due to magnetic interference from the vehicle’s thrusters and, potentially, the metallic structure of the culvert. As a result, the system has no absolute heading information and relies on gyroscope cues for estimating rotational motion and stabilizing the vehicle within the pipe.
\end{itemize}

\textbf{Environmental challenges} arise from the complex and imperfect nature of real-world culverts. In practice, a culvert may deviate significantly from the idealized cylindrical model. Its cross-section might not be perfectly circular and can exhibit corrugations or other surface irregularities that result in non-smooth internal geometries. Additionally, many culverts are not perfectly straight and may include bends, joints, or sections with noticeable deformation due to aging, sediment buildup, or structural damage. These variations introduce additional uncertainty into both the robot’s navigation and the interpretation of sensor data, as sonar returns may not correspond to a uniform or predictable surface.

\textbf{Sensing and localization challenges} stem from the robot’s limited and noisy sensing capabilities. At any given moment, the ROV receives only a raw range–intensity profile from the 360° sonar—which must be processed to extract a usable range measurement—a single processed range measurement from the Ping1D sonar, and depth and inertial data from the pressure sensor and IMU. Both sonars are highly susceptible to noise from acoustic reflections, multipath interference, and the confined geometry of the culvert, all of which degrade the reliability of range estimates. Moreover, because the 360° sonar requires approximately 4.5 s to complete a full rotation, the system must infer the vehicle’s position relative to the pipe cross-section from sparse, noisy, and temporally distributed measurements.

\section{Method}
This section outlines our approach for autonomous in-pipe traversal. 
We begin by processing the sonar profiles from the 360° rotating single-beam sonar to extract wall-range measurements. 
Next, we compute the pipe’s cross-sectional center from two sonar points—the downward-facing and rotating sonar returns—using a geometric construction. 
We then estimate the uncertainty of this center and compute the control commands accordingly. 
Finally, we regulate the robot’s translational and rotational motion with PD controllers to ensure safe and stable operation.

\begin{figure}[t]
    \centering
    \includegraphics[width=\linewidth]{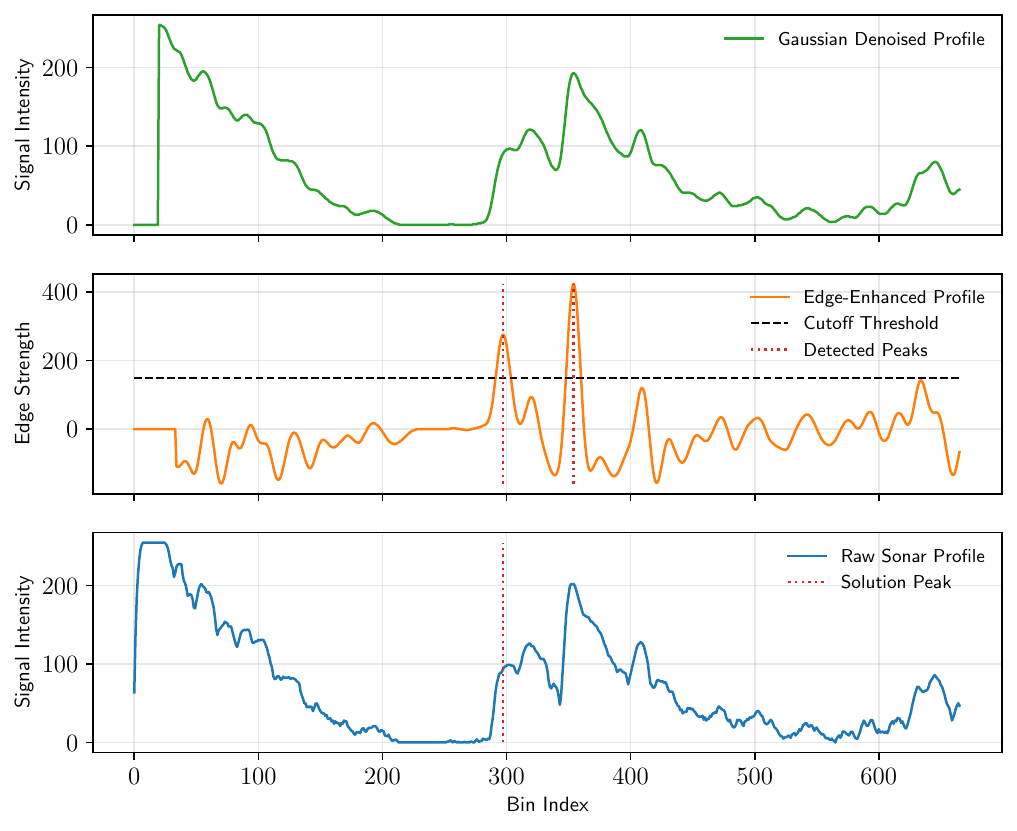}
    \caption{Sonar processing pipeline for range estimation. Sonar intensity profiles are processed through near-field suppression, denoising, and edge enhancement. The first significant return is detected as the pipe wall echo, converted to a range value, and smoothed over time to produce the final distance estimates for real-time navigation.}
    \label{fig:sonar}
\vspace{-2.5mm}
\end{figure}

\subsection{Sonar Range Estimation}

The goal of range estimation is to identify the first valid acoustic return from the pipe wall while rejecting near-field artifacts, multipath reflections, and noise. The procedure used to extract range measurements from each sonar intensity profile is described below and illustrated in Figure~\ref{fig:sonar}.

\paragraph*{1) Input and Near-Field Suppression}
Each sonar profile yields an intensity vector $\mathbf{s} = (s_1, s_2, \dots, s_N)$, where $s_i$ denotes the signal strength at the $i$-th range bin. The initial samples, often corrupted by transducer ring-down and near-field saturation, are removed by discarding the first $n_0$ bins, yielding $\tilde{\mathbf{s}} = (s_{n_0+1}, \dots, s_N)$, where $n_0$ defines the minimum valid detection range.

\paragraph*{2) Denoising and Edge Enhancement}
The truncated signal $\tilde{\mathbf{s}}$ is first smoothed using a one-dimensional Gaussian kernel $\mathcal{N}_\sigma$, producing $\mathbf{g} = \tilde{\mathbf{s}} * \mathcal{N}_\sigma$, where $*$ denotes discrete convolution and $\mathcal{N}_\sigma$ has standard deviation $\sigma$. This step reduces high-frequency noise while preserving gradual intensity transitions associated with valid echoes. To highlight the rising edge of the pipe-wall reflection, the smoothed profile is convolved with a symmetric kernel $\mathcal{W}$ of length $L$, i.e., $\mathbf{d} = \mathbf{g} * \mathcal{W}$, where $\mathcal{W}$ linearly tapers from $+1$ to $-1$.

\paragraph*{3) Peak Detection and Selection}
The peak detection procedure follows the approach described by Du~\textit{et al.}~\cite{DuKL06}. Candidate peaks are identified as local maxima in $\mathbf{d}$ that exceed a threshold $\tau$. The first valid peak $k^\star$ (the smallest index above $\tau$) is selected as the most probable wall return. If no valid peak is detected, the corresponding profile is discarded, and processing continues with the next profile.

\paragraph*{4) Range Conversion}
The corresponding range bin in the original signal is $b^\star = n_0 + k^\star + (L - 1)$, and the estimated range is computed as $\hat{r} = \tfrac{b^\star}{N - 1}\, r_{\max}$, where $r_{\max}$ is the maximum measurable range determined by the sonar sampling interval and the speed of sound in water.

\paragraph*{5) Temporal Smoothing}
Successive range estimates $\hat{r}_t$ are filtered for temporal consistency using a scalar Kalman filter~\cite{Barfoot17}, producing smooth, noise-resilient range estimates suitable for real-time navigation.

\smallskip
\noindent
\textbf{Summary:}
The proposed method—comprising near-field suppression, Gaussian denoising, edge enhancement, peak detection, range conversion, and temporal smoothing—extracts the first true echo in each sonar profile. This yields compact, noise-resilient range measurements that form the geometric foundation for pipe-center estimation and navigation control.

\subsection{Geometric Centers from Two Sonar Points}

Inside the circular pipe, the key geometric quantity for navigation and control is the robot’s displacement from the pipe centerline. Given the known pipe radius~$r$ and any two wall points measured by the sonars, there are \emph{two} circle centers consistent with the geometry, only one of which corresponds to the true cross-sectional center $\mathbf{c} = (x_0, z_0)$.

All coordinates are expressed in the robot’s body-fixed reference frame, whose origin is coincident with the onboard camera in the nose dome. The axes $(X_r, Y_r, Z_r)$ correspond to the lateral, forward, and vertical directions, respectively. The cross-sectional plane of interest lies in $(X_r, Z_r)$, which defines the plane used for center estimation.

The distances measured by the two sonars are first used to compute their respective wall-intersection points in each sonar’s local frame, based on the beam direction and the detected range. These points are then transformed into the robot’s reference frame using the known extrinsic calibration between each sonar and the robot body. The resulting intersection coordinates are
\[
\mathbf{p}_{\mathrm{D}} = (x_{\mathrm{D}},\, z_{\mathrm{D}}), 
\qquad
\mathbf{p}_{360} = (x_{360},\, z_{360}),
\]
where $\mathbf{p}_{\mathrm{D}}$ corresponds to the downward-facing sonar return and $\mathbf{p}_{360}$ to the 360° sonar return at an azimuth angle.

\begin{figure}[t]
    \centering
    \includegraphics[width=\linewidth]{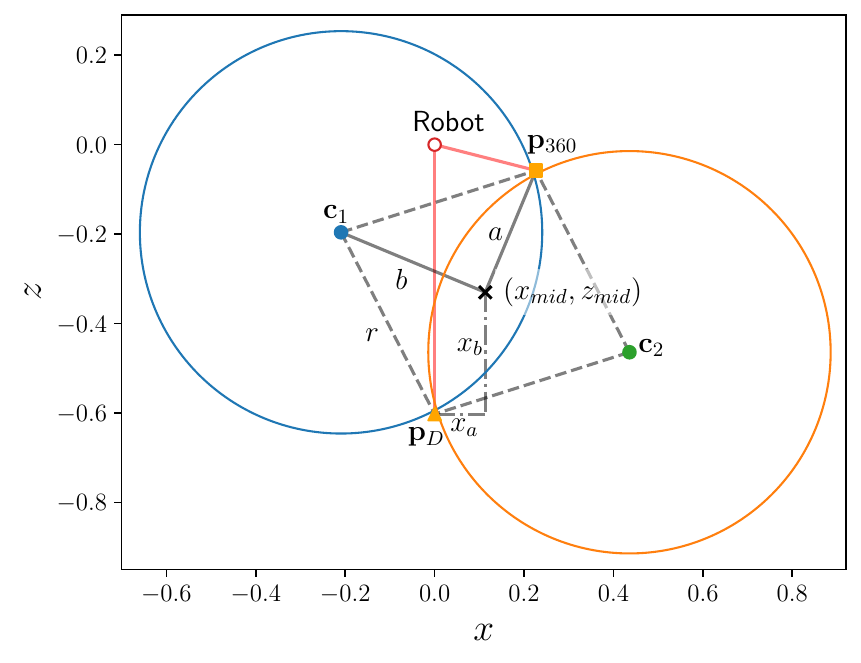}
    \caption{Geometric construction of candidate centers. Two sonar returns, $\mathbf{p}_{\mathrm{D}}$ and $\mathbf{p}_{360}$, define a chord on the circular pipe wall. The chord midpoint is computed, and two candidate circle centers ($\mathbf{c}_1$, $\mathbf{c}_2$) are obtained along the perpendicular bisector. The algorithm selects the physically consistent center based on geometric and temporal criteria.}
    \label{fig:solution}
\end{figure}

The problem is then framed as the classical geometry question: given two points on a circle of radius~$r$, find the circle’s center~\cite{David17}.  This process is illustrated in Fig.~\ref{fig:solution}, and proceeds as follows.

\paragraph*{Step 1: Chord Midpoint}
The two sonar points $\mathbf{p}_{\mathrm{D}}$ and $\mathbf{p}_{360}$ define a chord of the circular wall: 
\begin{equation}
x_a = \tfrac{1}{2}(x_{360} - x_{\mathrm{D}}), \qquad
z_a = \tfrac{1}{2}(z_{360} - z_{\mathrm{D}})\,.
\end{equation}

The two candidate centers, denoted $\mathbf{c}_1$ and $\mathbf{c}_2$, are symmetrically located along the perpendicular bisector of this chord. The chord midpoint is computed as
\begin{equation}
x_{\mathrm{mid}} = x_{\mathrm{D}} + x_a, \qquad
z_{\mathrm{mid}} = z_{\mathrm{D}} + z_a.
\end{equation}

\paragraph*{Step 2: Chord Length and Perpendicular Offset}
The half–chord length is $a = \sqrt{x_a^2 + z_a^2}$.
The perpendicular distance from the chord midpoint to each possible center is then $b = \sqrt{r^2 - a^2}$.

\paragraph*{Step 3: Centers by Perpendicular Offset}
The chord direction is $(x_a, z_a)$; a perpendicular direction is $(z_a, -x_a)$. Since both have length $a$, the unit perpendicular is $\left(\tfrac{z_a}{a},\, -\tfrac{x_a}{a}\right)$. 
Scaling by $b$ yields the two candidate centers: 
\begin{equation}
\label{eq:centers}
\begin{aligned}
\mathbf{c}_1 &= \left(x_{\text{mid}} + b \tfrac{z_a}{a},\;\; z_{\text{mid}} - b \tfrac{x_a}{a}\right),\\
\mathbf{c}_2 &= \left(x_{\text{mid}} - b \tfrac{z_a}{a},\;\; z_{\text{mid}} + b \tfrac{x_a}{a}\right).
\end{aligned}
\end{equation}

\paragraph*{Step 4: Selecting the Physically Consistent Center}
While both $\mathbf{c}_1$ and $\mathbf{c}_2$ satisfy the geometric constraints, only one is physically consistent with the robot’s position. Let $\hat{\mathbf{c}}_{k-1}$ denote the previous estimate of the pipe center. The selection proceeds as:

\begin{enumerate}
\item[a)] \textit{Inside--outside test:}
Select the center $\mathbf{c}_i$ for which the robot lies inside the circle of radius $r$ centered at $\mathbf{c}_i$:
\(
\|\mathbf{c}_i\| < r.
\)
If exactly one candidate passes, it is selected.

\item[b)] \textit{Proximity consistency:}
If both centers pass or both fail, select the candidate closest to the previous estimate:
\(
\mathbf{c}^\star = \arg\min_{\mathbf{c}_i\in\{\mathbf{c}_1,\mathbf{c}_2\}}\|\mathbf{c}_i-\hat{\mathbf{c}}_{k-1}\|.
\)

\begin{figure*}[ht]
   \centering
    \begin{subfigure}{0.32\linewidth}
        \includegraphics[width=\textwidth]{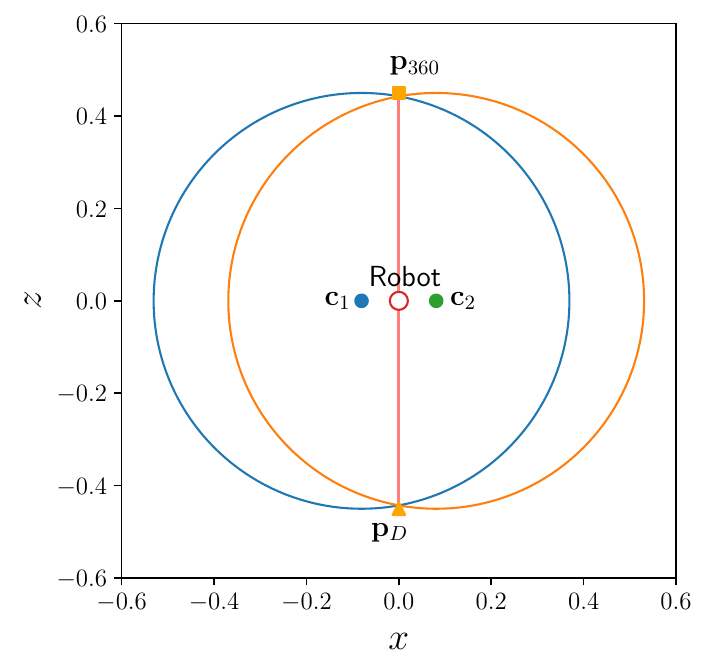}
        \caption{$180^{\circ}$ separation}
    \end{subfigure}
    \hfill
    \begin{subfigure}{0.32\linewidth}
        \includegraphics[width=\textwidth]{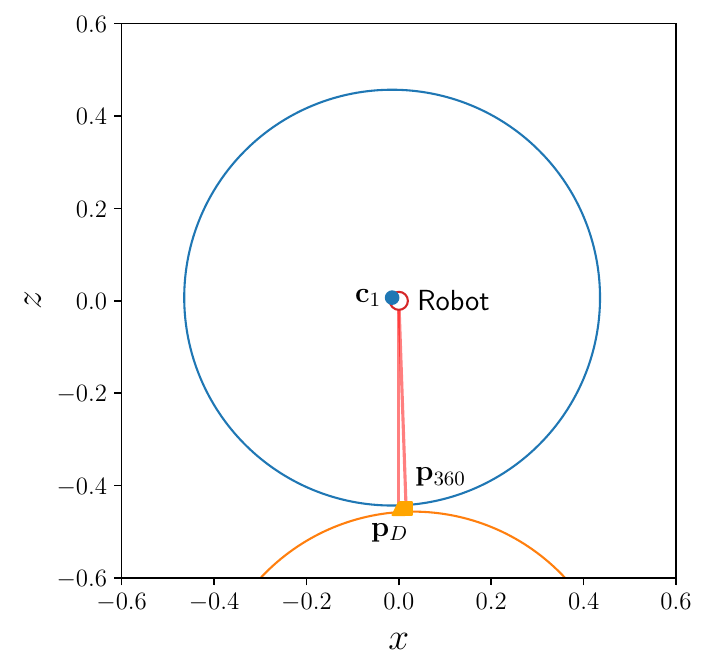}
        \caption{$0^{\circ}$ separation}
    \end{subfigure}
    \hfill
    \begin{subfigure}{0.32\linewidth}
        \includegraphics[width=\textwidth]{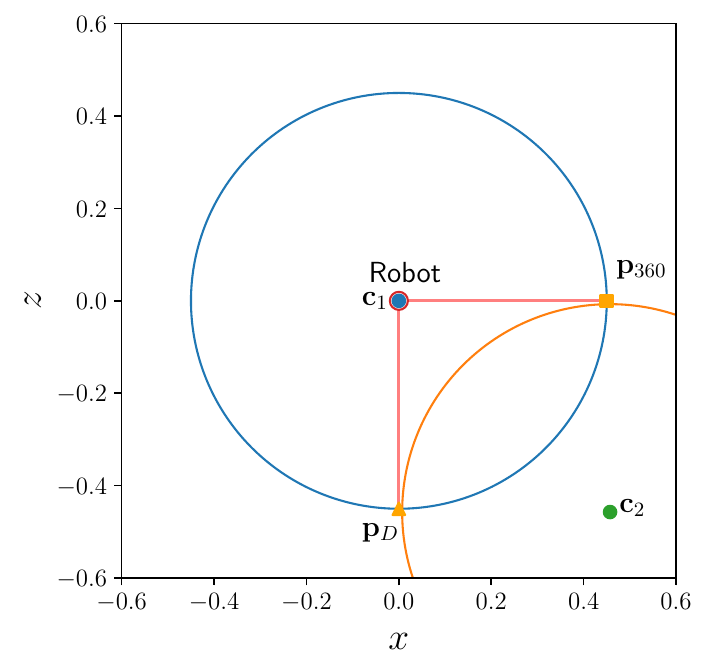}
        \caption{$90^{\circ}$ separation}
    \end{subfigure}
    \caption{Illustration of different angular separations between sonar beams: (a) $180^\circ$ separation (opposite directions) — ill-conditioned geometry; (b) $0^\circ$ separation (colinear beams) — degenerate intersection; (c) $90^\circ$ separation (perpendicular beams) — strong geometric constraint. Accurate centering is achieved when beams are approximately orthogonal.}
    \label{fig:stability}
\vspace{-2.5mm}
\end{figure*}

\item[c)] \textit{Temporal smoothing:}
To ensure temporal consistency and suppress high-frequency fluctuations in the estimated pipe center, consecutive center estimates are filtered using a discrete Kalman filter~\cite{Barfoot17}. At each update step, the predicted center position from the previous estimate is corrected based on the newly computed measurement $\mathbf{c}^\star$. The filter produces a smooth, noise-resilient trajectory of the pipe center suitable for real-time navigation and control.

\end{enumerate}

\subsection{Adaptive Measurement Covariance from Beam Alignment}

The accuracy of the estimated pipe center strongly depends on the angular separation between the two sonar beams, $\mathbf{p}_{\mathrm{D}}$ and $\mathbf{p}_{360}$. When these points are nearly colinear, their intersection provides poor lateral constraint; when they are close to orthogonal, the geometry is well conditioned and center estimation becomes more accurate (Fig.~\ref{fig:stability}). 

To capture how measurement reliability varies with beam alignment, the variances along the $X$- and $Z$-axes are modeled as functions of the beam separation angle $\theta = \angle(\mathbf{p}_{\mathrm{D}}, \mathbf{0}, \mathbf{p}_{360})$. This angle represents the separation between the two sonar measurement vectors, measured at the \emph{origin} in the robot frame. Each variance is represented as a mixture of three Gaussian components centered at $0^\circ$, $180^\circ$, and $360^\circ$:
\begin{equation}
\begin{aligned}
\sigma^2(\theta) &= A \!\left[
e^{-\frac{(\theta - 0^\circ)^2}{2w_{1}^2}} +
e^{-\frac{(\theta - 180^\circ)^2}{2w_{2}^2}} +
e^{-\frac{(\theta - 360^\circ)^2}{2w_{2}^2}}
\right].
\end{aligned}
\end{equation}

Here, $A$ scales the overall magnitude of the measurement variances, while $w_{i}$ ($i \in \{1,2\}$) define the width of each Gaussian component, determining how rapidly confidence changes with angular misalignment. This Gaussian mixture representation captures the periodic nature of the sonar geometry and ensures that variance peaks at colinear configurations ($\theta \approx 0^\circ$, $180^\circ$), and decreases near orthogonal alignments ($\theta \approx 90^\circ$ or $270^\circ$), where the geometric constraint is strongest. 

\begin{figure}[t]
    \centering
    \includegraphics[width=0.8\linewidth]{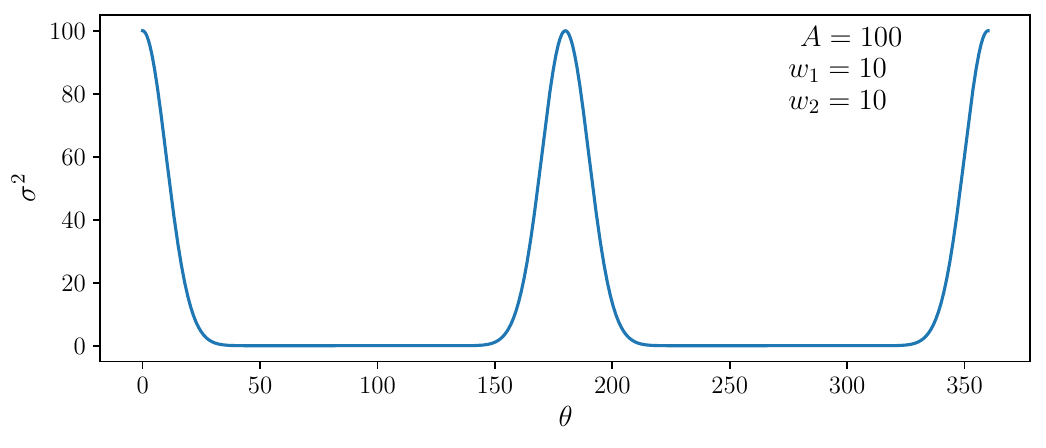}
    \caption{Example adaptive measurement covariance as a function of beam-separation angle $\theta$. Plot of the modeled measurement variance $\sigma^2(\theta)$ showing reduced uncertainty near orthogonal alignment and increased uncertainty when beams are nearly colinear. The covariance function governs adaptive weighting in the Kalman filter.}
    \label{fig:covariance}
\end{figure}

This adaptive variance function is used in the Kalman filter to prevent the estimator from becoming overconfident when the sonar returns are poorly aligned and to increase confidence when they are well separated. This improves filter stability and maintains robust center estimates under dynamically changing geometric conditions. Figure~\ref{fig:covariance} illustrates an example of the resulting angular variance profiles.

\subsection{Uncertainty-Aware Control Scaling}

The estimator provides the robot’s displacement from the pipe center $\hat{\mathbf{e}}_c = (\,\hat{e}_x,\, \hat{e}_z\,)$. The associated measurement covariances along each axis $\sigma_x^2$ and $\sigma_y^2$ used in the Kalman filter at time step~$k$ reflect the reliability of the sonar-based geometric measurements—larger values indicate higher uncertainty. To adapt the controller commands to this reliability, per-axis confidence weights are defined as
\begin{equation}
w_x = \frac{1}{1 + \sigma_x^2}, \qquad
w_z = \frac{1}{1 + \sigma_z^2}.
\label{eq:conf-weights}
\end{equation}

The weights $(w_x, w_z) \in (0,1]$ are applied in the linear PID controllers (Sec.~\textit{Linear Control Commands}) to scale the motor thrusts.  
Low-confidence sonar readings (large~$\sigma^2$) yield smaller weights, reducing control effort and preventing overreaction to unreliable measurements.  
Conversely, when sonar estimates are precise (small~$\sigma^2$), the weights approach unity, restoring full magnitude of the control commands.

\subsection{Linear Control Commands}

The linear control subsystem regulates motion along the lateral ($X_r$), vertical ($Z_r$), and forward ($Y_r$) axes in the robot body frame.  
Each translational axis is controlled by a proportional–derivative (PD) controller~\cite{LynchP17} that minimizes the estimated displacement from the pipe center.  
The controller output is scaled by confidence weights $(w_x, w_z)$, reducing control effort when sonar uncertainty is high and ensuring smooth, stable motion.

\paragraph*{1: Lateral Control}
\begin{equation}
u_x = \bigl(K_{Px}\,\hat{e}_x + K_{Dx}\,\dot{\hat{e}}_x\bigr)\,w_x,
\end{equation}
where $\hat{e}_x$ is the estimated lateral displacement from the pipe center and $\dot{\hat{e}}_x$ its time derivative.  
The proportional gain $K_{Px}$ corrects position error, while the derivative gain $K_{Dx}$ damps oscillations by opposing lateral velocity.  
The confidence weight $w_x$ scales the control magnitude to reflect sonar reliability.

\paragraph*{2: Vertical Control (Depth-Hold PD)}
\begin{equation}
h^\star =
\begin{cases}
h + \hat{e}_z, & w_z > \tau,\\[4pt]
h^\star, & \text{otherwise,}
\end{cases}
\end{equation}
\begin{equation}
u_z = K_{Pz}\,(h^\star - h) + K_{Dz}\,(\dot{h}^\star - \dot{h}),
\end{equation}
where $\hat{e}_z$ is the estimated vertical offset from the pipe center, and $h$ is the measured depth from the pressure sensor.  
The target altitude $h^\star$ is updated by the sonar estimate only when the confidence $w_z$ exceeds the threshold $\tau$.
This gating ensures smooth vertical motion and robustness under intermittent sonar measurements.

\paragraph*{3: Forward Control}
\begin{equation}
u_y =
\begin{cases}
u_{\mathrm{forward}}, & \|\hat{\mathbf{e}}_c - (0,\, h_{\mathrm{trg}})\| < \varepsilon,\\[4pt]
0, & \text{otherwise,}
\end{cases}
\end{equation}
where $u_{\mathrm{forward}}$ is a constant assist thrust applied along $Y_r$.  
Forward motion is enabled only when the estimated cross-sectional error $\hat{\mathbf{e}}_c$ lies within a tolerance $\varepsilon$ of the desired offset from the pipe center, $(0,\, h_{\mathrm{trg}})$.  
For a point robot, $h_{\mathrm{trg}} = 0$; however, for a real vehicle, a small nonzero offset may be required to account for its physical dimensions.
This gating prevents forward thrust when the robot is significantly misaligned, reducing the risk of wall contact.

\subsection{Rotational Control Commands}

The rotational control subsystem stabilizes the vehicle’s attitude about roll ($\phi$), pitch ($\theta$), and yaw ($\psi$) in the body frame.  
Each axis is independently regulated by a PD controller that generates the corresponding control moment $(u_\phi, u_\theta, u_\psi)$ to maintain a stable and level orientation.

\paragraph*{1: Roll and Pitch}
\begin{equation}
u_\phi = -K_{P\phi}\,\phi - K_{D\phi}\,\dot{\phi}, \qquad
u_\theta = -K_{P\theta}\,\theta - K_{D\theta}\,\dot{\theta},
\end{equation}
where $K_{P\phi}$ and $K_{P\theta}$ are proportional gains for roll and pitch, and $K_{D\phi}$, $K_{D\theta}$ are derivative gains that damp angular rates.  
These controllers maintain a level attitude and ensure the sonar sensors remain correctly oriented.

\paragraph*{2: Yaw (Gyro-Based Heading Hold)}
\begin{equation}
\hat{\psi}(t_k) = \hat{\psi}(t_{k-1}) + \dot{\psi}_{\mathrm{gyro}}(t_k)\,\Delta t, \qquad \hat{\psi}(t_0) = \psi_0,
\end{equation}
\begin{equation}
u_\psi = K_{P\psi}\,(\psi_0 - \hat{\psi}) - K_{D\psi}\,\dot{\psi}_{\mathrm{gyro}},
\end{equation}
where $\dot{\psi}_{\mathrm{gyro}}$ is the IMU-measured yaw rate and $\psi_0$ the initial heading.  
The controller maintains the desired yaw orientation using gyroscopic feedback alone, eliminating the need for a magnetometer and providing stable heading control even in magnetically disturbed environments.

\smallskip
\noindent
\textbf{Summary:}
The complete control system provides six-degrees-of-freedom regulation through body-frame thrusts $(u_x, u_y, u_z)$ and rotational moments $(u_\phi, u_\theta, u_\psi)$.  
Confidence weights $(w_x, w_z)$ adaptively modulate translational actuation based on sonar reliability, while PD-based attitude stabilization ensures consistent sensor alignment for robust geometric centering and safe pipe traversal.
\section{Experiments}

This section presents the experiments conducted to validate the proposed approach. First, we evaluate the sonar range extraction method by comparing the estimated wall-range measurements to manually labeled ground-truth distances from real-world sonar data. Next, we assess the accuracy of the pipe-center estimation and control strategy in a simulated 2D environment. Finally, we demonstrate the complete system operating on a real underwater robot.

\subsection{Sonar Range Estimation}

\begin{figure}[t]
    \centering
    \includegraphics[width=\linewidth]{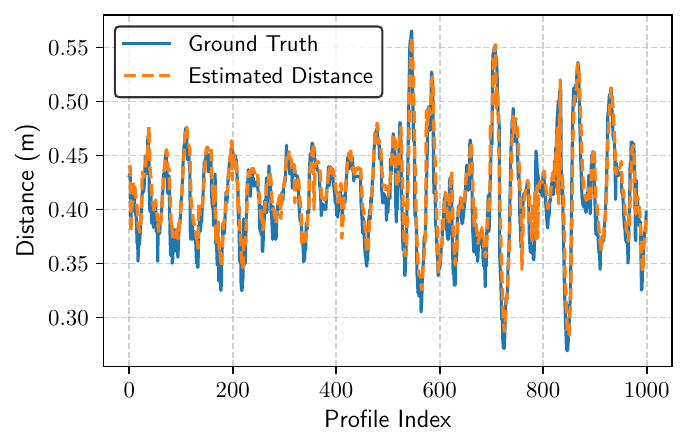}
    \caption{
        Comparison between the ground-truth wall distances (solid blue) and the estimated wall-range measurements (dashed orange) across sonar intensity profiles. 
     The method reliably identifies the first wall return, producing range estimates that closely match ground truth, with a root mean square error~(RMSE) of 0.02~m.}
    \label{fig:sonar_benchmark}
\end{figure}

We validated the sonar range estimation method using real-world sonar profiles collected with a Blue Robotics Ping360 sonar mounted on a BlueROV2 vehicle. 
Data was acquired by maneuvering the vehicle through a flexible fabric pipe with an inner diameter of 0.46~cm, submerged in a swimming pool to form a straight tunnel.
A total of 1000 sonar profiles were manually labeled with the ground-truth wall ranges based on visual interpretation of the sonar intensity profiles. 
Figure~\ref{fig:sonar_benchmark} compares the estimated ranges produced by the proposed sonar processing pipeline with the corresponding ground-truth labels. 
The method consistently produces wall-distance estimates that closely match ground truth, achieving a root mean square error~(RMSE) of 0.02~m across the dataset, demonstrating reliable sonar range estimation.

\subsection{Pipe Center Estimation and Tracking}

In this experiment, we implemented a 2D geometric simulation of a point robot operating inside a circle with a radius of 0.46~cm. The two sonar measurements were simulated by projecting rays from the robot center to the circular boundary—one directed downward and the other swept across azimuth angles from $0^\circ$ to $360^\circ$ in incremental steps. To emulate real-world sensor uncertainty, uniform noise between $-0.04$~m and $0.04$~m was added to the sonar intersection coordinates.

The proposed center estimation and tracking method was evaluated in this environment. Each simulation was initialized with the robot placed at a random location within the circle, and the experiment proceeded until the rotating sonar completed three full $360^\circ$ sweeps. The experiment was repeated 100 times to assess consistency and robustness.

\begin{figure}[t]
    \centering
    \includegraphics[width=0.95\linewidth]{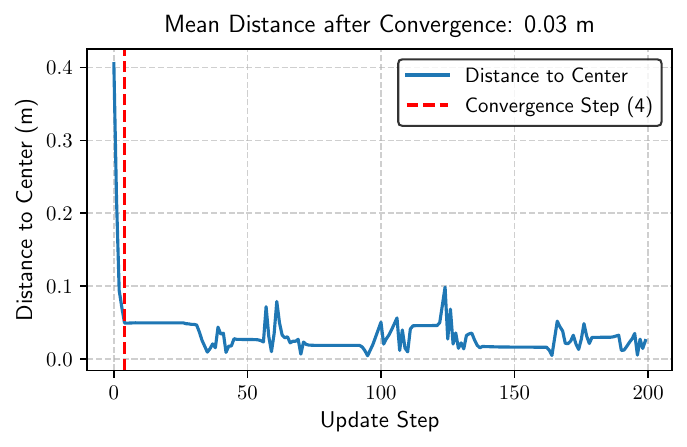}
    \caption{
        Convergence behavior of the simulated robot during pipe-center tracking. 
        The system reliably converges within a few update steps, after which the robot remains close to the center.
    }
    \label{fig:convergence}
\end{figure}

On average, the robot required 10 update steps to converge within 0.05~m of the center, with each step corresponding to a new pair of sonar measurements. After convergence, it maintained a mean steady-state error of approximately 0.029~m across all runs. Figure~\ref{fig:convergence} illustrates the convergence behavior in simulation, showing an example run where the robot’s distance to the center decreases over time.

\subsection{Field Experiment}

\begin{figure*}[ht]
   \centering
    \begin{subfigure}{0.24\linewidth}
        \includegraphics[width=0.99\textwidth]{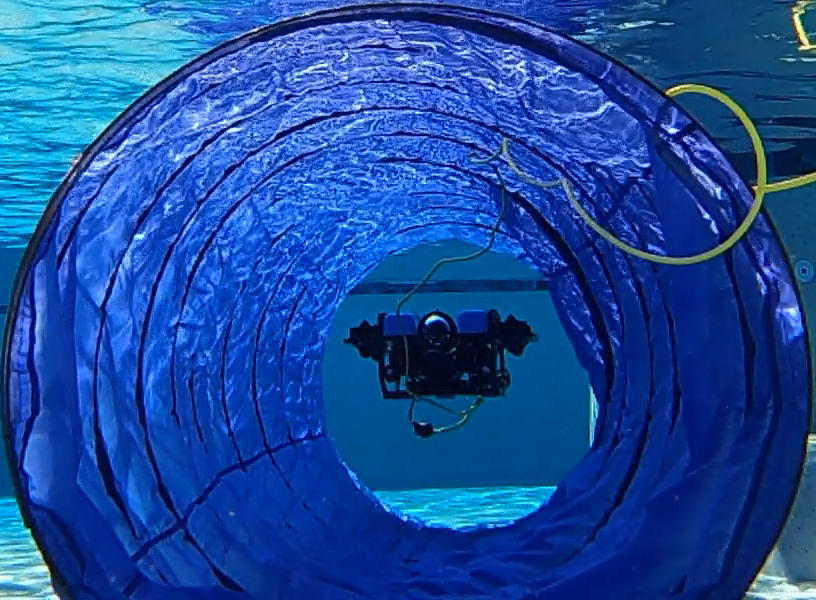}
    \end{subfigure}
    \hfill
    \begin{subfigure}{0.24\linewidth}
        \includegraphics[width=0.99\textwidth]{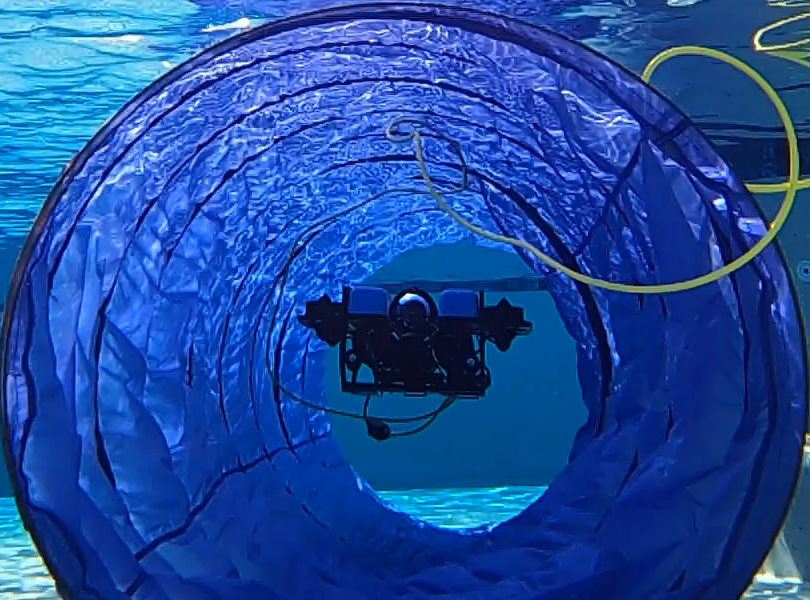}
    \end{subfigure}
    \hfill
    \begin{subfigure}{0.24\linewidth}
        \includegraphics[width=0.99\textwidth]{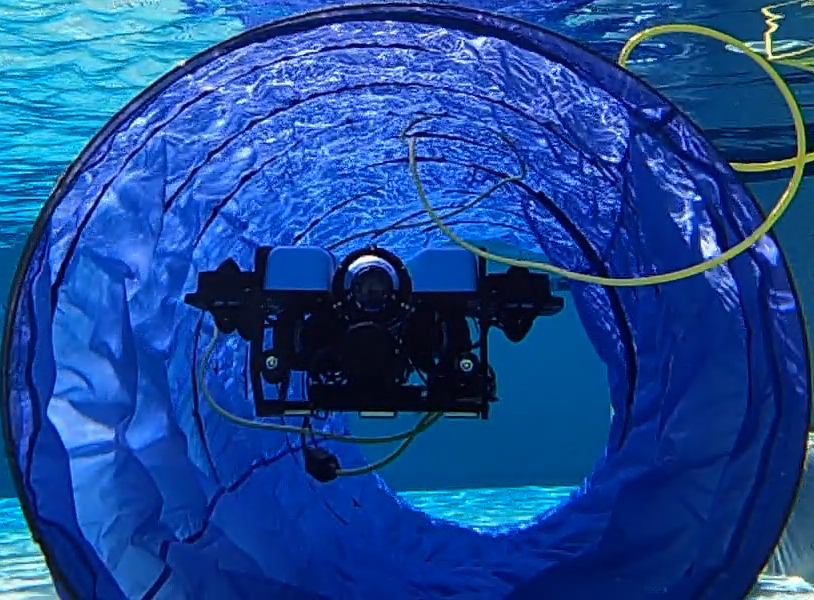}
    \end{subfigure}
    \hfill
    \begin{subfigure}{0.24\linewidth}
        \includegraphics[width=0.99\textwidth]{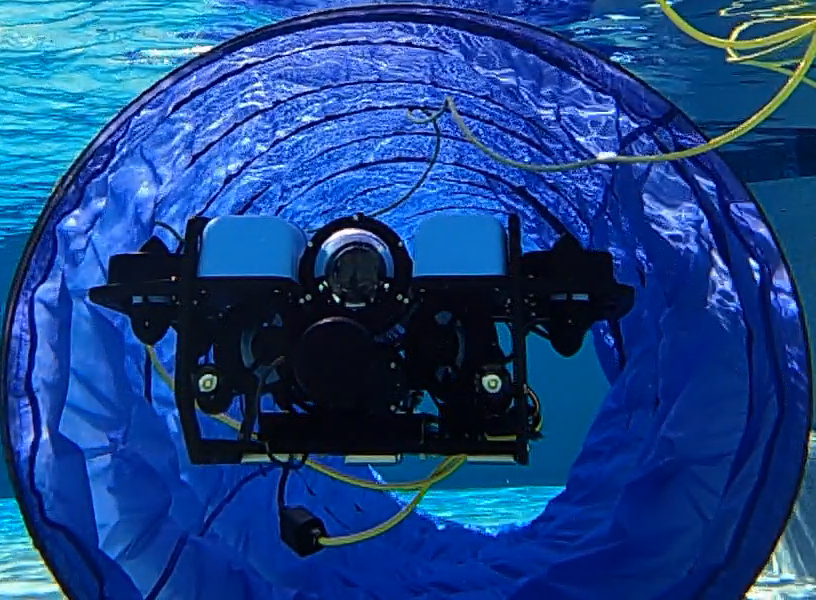}
    \end{subfigure}
    \caption{Representative images from full-length traversal experiments in the fabric culvert. The robot maintains good alignment throughout, despite deformations in the pipe.}
    \label{fig:traversal}
\end{figure*}

The proposed method was validated in a controlled aquatic environment designed to emulate a submerged culvert. A flexible fabric pipe with an inner diameter of 0.46~cm was submerged in a swimming pool to form a straight tunnel. In practice, the pipe was never perfectly straight; its flexible structure introduced small bends and twists along its length, further increasing the complexity of the navigation task. Continuous water movement from a nearby fountain caused the fabric structure to oscillate and shift laterally by several centimeters over time. During testing, the pipe also developed a localized deformation near its midpoint, producing an inward bend that partially reduced the cross-sectional area. These characteristics introduced realistic disturbances and boundary variations, providing a rigorous environment for evaluating the robustness of the proposed centering and traversal method.

Experiments were conducted using a BlueROV2 Heavy vehicle operating inside the submerged pipe. The robot was equipped with an IMU, a pressure sensor, and two sonars: a downward-facing Ping1D single-beam sonar providing direct range measurements, and a nose-mounted Ping360 mechanically scanning single-beam sonar for cross-sectional sensing. The Ping360 performed 360° scans in 9° increments, completing a full rotation approximately every 4.5 seconds and producing one intensity profile per azimuth angle. Before each trial, the vehicle was manually aligned with the pipe’s longitudinal axis.

To obtain range information from the Ping360 sonar, the sonar data processing method described in the methods section was applied in real time. The resulting range estimates were then used as geometric inputs for pipe-center estimation.

Two main experiments were conducted to evaluate the proposed approach, with corresponding experiment videos provided in the video appendix. Additional implementation details can be found in the code appendix.

\begin{enumerate}
    \item \textbf{Auto-Centering Validation:} 
    In this experiment, the robot’s ability to autonomously re-center itself at the entrance of the pipe without forward thrust was tested. The vehicle was manually displaced from the center and externally perturbed along the four cardinal directions ($\pm X$, $\pm Z$) and diagonally. The adaptive controller successfully recovered the robot to the centerline using only sonar, IMU, and pressure sensor feedback. Recovery trajectories were smooth and stable despite ambient water flow and minor pipe deformations. 

    \item \textbf{Pipe Traversal Trials:} 
    The second experiment evaluated autonomous forward traversal through the entire deformable fabric culvert. Multiple runs were performed from different initial offsets to test repeatability and robustness. During traversal, the robot encountered both gradual curvature along the pipe and a mid-pipe constriction caused by local fabric deformation. The system maintained stable centering, successfully navigating through the deformed and curved sections. Figure~\ref{fig:traversal} shows representative images from this experiment. 

\end{enumerate}
\section{Conclusion}

This paper presented a two-point, sonar-based framework for autonomous centering and traversal of submerged pipelines using a minimal sensor suite. We developed a computationally efficient algorithm for extracting wall-range measurements from raw single-beam sonar intensity data, providing reliable wall detection under noise and multipath interference while maintaining real-time performance on embedded hardware. The pipe-centering method combines a closed-form geometric formulation with adaptive confidence weighting, enabling a free-swimming underwater robot to estimate its displacement from the pipe center and maintain stable alignment during motion.

The proposed system was validated through three complementary experiments. First, the sonar range extraction method achieved accurate wall-range estimation on real-world data, with a root mean square error~(RMSE) of 0.02~m. Second, a 2D geometric simulation demonstrated convergence to within 0.05~m of the true pipe center in fewer than ten update steps on average, confirming the accuracy and stability of the estimation and control framework. Finally, full-scale field experiments in a submerged, deformable fabric culvert showed that the BlueROV2 vehicle maintained stable centering and successfully traversed sections with curvature, deformation, and ambient flow disturbances. Together, these results demonstrate that stable in-pipe navigation can be achieved using lightweight acoustic sensing and efficient geometric reasoning.

Despite its strong performance, the approach has several limitations. First, it assumes the robot begins approximately aligned with the pipe axis; large yaw misalignments can reduce estimation stability. Second, it assumes a circular pipe cross-section with a known radius. Third, the system depends on several parameters—such as filter gains, sonar thresholds, and confidence-weighting factors—that must be carefully tuned. Future work may reduce this tuning burden through adaptive or learning-based optimization methods~\cite{IqbalKJRJ22}.

Beyond these improvements, future efforts will focus on extending the approach to more complex pipe networks and enhancing robustness under stronger flow and turbulence conditions.

\bibliography{references}

@ARTICLE{TanSAWF19,
  author={Tan, Chee How and Shaiful, Danial Sufiyan bin and Ang, Wei Jun and Win, Shane Kyi Hla and Foong, Shaohui},
  journal={IEEE Robotics and Automation Letters}, 
  title={Design Optimization of Sparse Sensing Array for Extended Aerial Robot Navigation in Deep Hazardous Tunnels}, 
  year={2019},
  volume={4},
  number={2},
  pages={862-869},
}

@article{MonteroVMJB15,
title = {Past, present and future of robotic tunnel inspection},
journal = {Automation in Construction},
volume = {59},
pages = {99-112},
year = {2015},
issn = {0926-5805},
author = {R. Montero and J.G. Victores and S. Martínez and A. Jardón and C. Balaguer},
}

@article{TardioliRSRLVM19,
author = {Tardioli, Danilo and Riazuelo, Luis and Sicignano, Domenico and Rizzo, Carlos and Lera, Francisco and Villarroel, José L. and Montano, Luis},
title = {Ground robotics in tunnels: Keys and lessons learned after 10 years of research and experiments},
journal = {Journal of Field Robotics},
volume = {36},
number = {6},
pages = {1074-1101},
year = {2019}
}

@INPROCEEDINGS{MateosDV13,
  author={Mateos, Luis A. and Dominguez, Marcos Rodriguez and Vincze, Markus},
  booktitle={2013 IEEE/RSJ International Conference on Intelligent Robots and Systems}, 
  title={Automatic in-pipe robot centering from 3D to 2D controller simplification}, 
  year={2013},
  volume={},
  number={},
  pages={258-265},
}

@INPROCEEDINGS{KakogawaSH14,
  author={Kakogawa, Atsushi and Ma, Shugen and Hirose, Shigeo},
  booktitle={2014 IEEE International Conference on Robotics and Automation (ICRA)}, 
  title={An in-pipe robot with underactuated parallelogram crawler modules}, 
  year={2014},
  volume={},
  number={},
  pages={1687-1692},
}

@INPROCEEDINGS{ChenF23,
  author={Chen, Heping and Fang, Siwen},
  booktitle={2023 IEEE 13th International Conference on CYBER Technology in Automation, Control, and Intelligent Systems (CYBER)}, 
  title={Optimal Piping Robot Design for Inner Pipe Operations}, 
  year={2023},
  volume={},
  number={},
  pages={241-245},
}

@Article{PatelAATOASAAFAHJS24,
author={Patel, Sahejad
and Abdellatif, Fadl
and Alsheikh, Mohammed
and Trigui, Hassane
and Outa, Ali
and Amer, Ayman
and Sarraj, Mohammed
and Al Brahim, Ahmed
and Alnumay, Yazeed
and Felemban, Amjad
and Alrasheed, Ali
and Halawani, Abdulwahab
and Jifri, Hesham
and Jaleel, Hassan
and Shamma, Jeff},
title={Multi-robot system for inspection of underwater pipelines in shallow waters},
journal={International Journal of Intelligent Robotics and Applications},
year={2024},
month={Mar},
day={01},
volume={8},
number={1},
pages={14-38},
}

@article{SewellPS24,
  title={Non-linear AUV controller design using logic-based switching PID Control},
  author={Sewell, Tazden and Padayachee, Jared and Snedden, Glen C},
  journal={International Journal of Mechanical Engineering and Robotics Research},
  volume={13},
  number={2},
  pages={227--240},
  year={2024}
}

@ARTICLE{KazeminasabSSJRZB21,
  author={Kazeminasab, Saber and Sadeghi, Neda and Janfaza, Vahid and Razavi, Moein and Ziyadidegan, Samira and Banks, M. Katherine},
  journal={IEEE Access}, 
  title={{Localization, Mapping, Navigation, and Inspection Methods in In-Pipe Robots: A Review}}, 
  year={2021},
  volume={9},
  number={},
  pages={162035-162058},
}

@INPROCEEDINGS{WuNKYB15,
  author={Wu, You and Noel, Antoine and Kim, David Donghyun and Youcef-Toumi, Kamal and Ben-Mansour, Rached},
  booktitle={2015 IEEE/RSJ International Conference on Intelligent Robots and Systems (IROS)}, 
  title={Design of a maneuverable swimming robot for in-pipe missions}, 
  year={2015},
  volume={},
  number={},
  pages={4864-4871},
}

@INPROCEEDINGS{MazumdarLFH12,
  author={Mazumdar, Anirban and Lozano, Martin and Fittery, Aaron and Harry Asada, H.},
  booktitle={2012 IEEE International Conference on Robotics and Automation}, 
  title={A compact, maneuverable, underwater robot for direct inspection of nuclear power piping systems}, 
  year={2012},
  volume={},
  number={},
  pages={2818-2823},
}

@article{ManjunathaSGM18,
title = {A Low Cost Underwater Robot with Grippers for Visual Inspection of External Pipeline Surface},
journal = {Procedia Computer Science},
volume = {133},
pages = {108-115},
year = {2018},
note = {International Conference on Robotics and Smart Manufacturing (RoSMa2018)},
author = {M. Manjunatha and A. Arockia Selvakumar and Vivek P Godeswar and R. Manimaran},
}

@INPROCEEDINGS{ShiCGGHPXST17,
  author={Shi, Liwei and Chen, Zhan and Guo, Shuxiang and Guo, Ping and He, Yanlin and Pan, Shaowu and Xing, Huiming and Su, Shuxiang and Tang, Kun},
  booktitle={2017 IEEE International Conference on Mechatronics and Automation (ICMA)}, 
  title={An underwater pipeline tracking system for amphibious spherical robots}, 
  year={2017},
  volume={},
  number={},
  pages={1390-1395},
}

@INPROCEEDINGS{NarimaniNL09,
  author={Narimani, Mehdi and Nazem, Soroosh and Loueipour, Mehdi},
  booktitle={OCEANS 2009-EUROPE}, 
  title={Robotics vision-based system for an underwater pipeline and cable tracker}, 
  year={2009},
  volume={},
  number={},
  pages={1-6},
}

@MISC{David17,
    TITLE = {Finding the Center of a circle given two points and a radius (algebraically)},
    AUTHOR = {David K (https://math.stackexchange.com/users/139123/david-k)},
    HOWPUBLISHED = {Mathematics Stack Exchange},
    year={2017},
    URL = {https://math.stackexchange.com/q/1781546}
}

@article{TurG10,
author = {Mirats Tur, Josep M. and Garthwaite, William},
title = {Robotic devices for water main in-pipe inspection: A survey},
journal = {Journal of Field Robotics},
volume = {27},
number = {4},
pages = {491-508},
year = {2010}
}

@book{Barfoot17,
  title={State Estimation for Robotics},
  author={Barfoot, T.D.},
  isbn={9781107159396},
  year={2017},
  publisher={Cambridge University Press}
}

@book{LynchP17,
  title={Modern Robotics},
  author={Lynch, K.M. and Park, F.C.},
  isbn={9781107156302},
  year={2017},
  publisher={Cambridge University Press}
}

@InProceedings{YangLRL23,
author="Yang, Pengzhi
and Liu, Haowen
and Roznere, Monika
and Li, Alberto Quattrini",
editor="Billard, Aude
and Asfour, Tamim
and Khatib, Oussama",
title="Monocular Camera and Single-Beam Sonar-Based Underwater Collision-Free Navigation with Domain Randomization",
booktitle="Robotics Research",
year="2023",
publisher="Springer Nature Switzerland",
pages="85--101"}

@book{Nielsen91,
  title={Sonar Signal Processing},
  author={Nielsen, R.O.},
  isbn={9780890064535},
  series={Acoustics and Signal Processing Library},
  year={1991},
  publisher={Artech House}
}

@article{DuKL06,
    author = {Du, Pan and Kibbe, Warren A. and Lin, Simon M.},
    title = {Improved peak detection in mass spectrum by incorporating continuous wavelet transform-based pattern matching},
    journal = {Bioinformatics},
    volume = {22},
    number = {17},
    pages = {2059-2065},
    year = {2006},
    month = {07},
    issn = {1367-4803},
}

@inproceedings{IqbalKJRJ22,
author = {Iqbal, Md Shahriar and Krishna, Rahul and Javidian, Mohammad Ali and Ray, Baishakhi and Jamshidi, Pooyan},
title = {Unicorn: reasoning about configurable system performance through the lens of causality},
year = {2022},
isbn = {9781450391627},
publisher = {Association for Computing Machinery},
address = {New York, NY, USA},
booktitle = {Proceedings of the Seventeenth European Conference on Computer Systems},
pages = {199–217},
numpages = {19},
location = {Rennes, France},
series = {EuroSys '22}
}


\end{document}